# A Reputation System for Artificial Societies

Anton Kolonin[1,2], Ben Goertzel[2], Deborah Duong[2], Matt Ikle[2]

[1]Aigents Group, Novosibirsk, Russia
[2]SingularityNET Foundation, Amsterdam, Netherlands
{anton, ben, deborah, matt}@singularitynet.io

**Abstract.** One approach to achieving artificial general intelligence (AGI) is through the emergence of complex structures and dynamic properties arising from decentralized networks of interacting artificial intelligence (AI) agents. Understanding the principles of consensus in societies and finding ways to make consensus more reliable becomes critically important as connectivity and interaction speed increase in modern distributed systems of hybrid collective intelligences, which include both humans and computer systems. We propose a new form of reputation-based consensus with greater resistance to reputation gaming than current systems have. We discuss options for its implementation, and provide initial practical results.

## 1 Introduction and background

Since the appearance of decentralized and distributed computer systems without centralized governance, verification of the reputation of participants was well understood to be a problem, and has been studied in its many aspects [1]. A reliable solution for the determination of reputation has turned critical for peer-to-peer systems, where every node can communicate with every other node in the network [2]. The standard theoretical framework to approach such a solution comes from the so-called "Byzantine Generals Problem" where a variable number of participants with variable levels of trust vote independently in order to reach consensus in respect to a decision to be recorded in some sort of public ledger, so that the decision is known and beneficial to the entire community[3]. Since the trustworthiness level of every node in the system is not known in advance, there is a need to mitigate the risk of alien nodes trying to ally to take over the consensus in favor of the alien alliance over the remaining members. In existing distributed computation systems based on blockchain technology, different consensus algorithms implementing various forms of weighted voting are being implemented, each of which provides certain heuristics upon which quality of a node in the computer network can be used to infer its prospective level of trust [4].

A potentially viable route to artificial general intelligence (AGI) lies via emergent structures and dynamics in a decentralized network of interacting artificial intelli-gence (AI) agents., With a need to prevent various pathologies in such a network, a high quality reputation system is needed. Ensuring high quality reputation assignment requires AI, thus there is leading to a mutual recursion between AGI and reputation assessment in a decentralized AI networks. Thus, being able to build reliable reputation systems is critical for AGI problem solution, and any paradigm of AGI should address the "assignment of credit problem" in some way or another. This always ultimately involves A) simple base algorithm for credit assignment in simple common cases, and B) a framework for addressing assignment of credit via mutual recursion with other aspects of general intelligence for more complex cases.

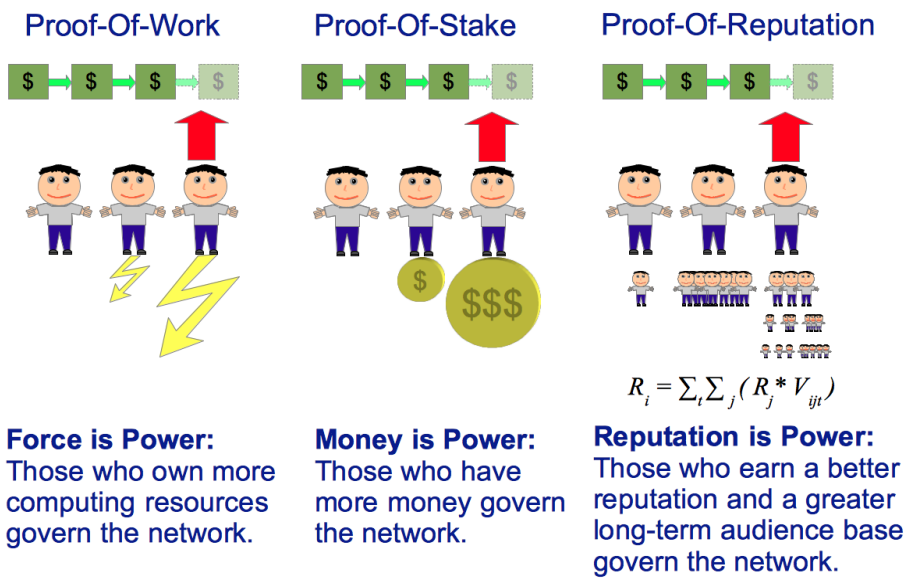

**Fig.1.** Types of consensus in distributed systems such.

Most of the consensus algorithms discussed in earlier works [4] and implemented in existing popular distributed computation systems such as Ethereum and Bitcoin are still vulnerable to takeover. The consensus algorithm called Proof-of-Work (POW), in which a member of its system votes by amount of computing power that the member possesses, can be abused by an alliance which temporarily concentrates more than 51% of computer power trying to reach consensus at the moment. From an historical perspective of human society, this algorithm corresponds to the "rule of power", specific to most ancient societies. Another known consensus algorithm called Proof-of-Stake (POS) implies voting by amount of financial funds each member has at stake. This is similar to consensus in modern capitalist communities with "rule of power". It leads to the phenomenon of extreme income disparity in which one member or class of member is able to take over the consensus. The advanced version of POS called Delegated-Proof-of-Stake attempts to fix this problem by explicit delegation of the right to rule to "delegated", appointed by members with greater stake, but this just makes the distributed system manually controlled.

In this work we develop an advanced version of a consensus algorithm called Proof-of-Reputation, which assumes "rule of reputation". In this case, the ability of a member to have an impact on consensus can be identified by the amount of reputation, social capital or "karma" actually earned by the member in the course of interacting with other members in a given time frame, taking into account the reputations of these members themselves [5].

A side effect of the ability to compute a reliable social consensus in a community of members is the ability to identify the reliability of each of the members, enabling the most efficient and safest communications between them. This is suggested by our earlier work on design for reputation system for a decentralized marketplace of artificial intelligence services called SinglularityNET [6]. Proof-of-Reputation provides the ability to measure and track the reputation dynamics of every member in society This could be applied to any community of artificial agents, real humans interacting online, or even hybrid human-machine societies [7].

## 2    Model of computing reputations

The computational model described below is based on previous ideas of measuring quantitative properties of social graphs accordingly to earlier works [7,8,9] adapted to the problem of computing social reputations as grounds for building a better consensus algorithm [3,4,5,6].

The reputation $R_i(t)$ of society member $i$ at moment $t$ can be computed incrementally on the basis of its own reputation at the previous moment $R_i(t-1)$, and some default reputation $Rd$ taken as its initial reputation. Changes in the reputation of $i$ can be caused by different sorts of ratings issued by multiple other members $j$, in respect to a particular aspect of reputations $k$ and specific domain category of reputation $c$. The aspect $k$ is assumed to be a generic measure like reliability, quality or timeliness while the category $c$ may identify an area of a member's expertise such as painting, stock prediction or pizza delivery.

Ratings can be divided into two types. First, there are endorsing ratings $S_{ijkc}$, which may be or may not be be present at any time $t$, being granted or revoked from $j$ to $i$. Next, there are transactional ratings $F_{ijkce}$ that can be recorded in a history of interactions, being associated with either financial transactions from $j$ to $i$ (financial ratings) or acts of voting (voting ratings) in respect to particular events $e(t)$, such as publications, posts, comments, nominations or tasks and duties being served by $i$ in respect to $j$. Most ratings can be either explicit or implicit. Explicit voting ratings come with rank value expressed as positive, negative or any number at some scale, while implicit ones are comments and reviews authored by $j$ in respect to $i$ where actual value or rating should be somehow from the media used for comment or review, such as natural language text. Endorsing ratings $S_{ijkc}$ may be backed up with financial stake value $Q_{ij}$. Transactional voting ratings $F_{ijkce}$ can be backed up by a financial value $G_{ije}$. For exam-ple,, the value of a customer's $j$ vote in respect to the quality of service provider $i$ may be weighted with account of cost of the entire service $e(t)$.

The rating values maybe scaled in the range *-1* to *1* for negative and positive ratings, while for presentation purposes they may be scaled to *-5* to *5*, *0* to *10* or whatever seems visually intuitive. For financial ratings, experimentation with Ethereum blockchain, has shown that it is desirable to normalize the nonlinear distributions of financial values of transactions as follows:

$$F'_{ijce} = log_{10}(F'_{ijce})/MAX(log_{10}(F'_{ijce}))$$

Ratings for different aspects $k$ can be blended to infer overall reputation using a system-wide blending parameter $H_k$. Then, the following formulae can identify differential reputation at time $t_n$ as a relative increase of reputation due to endorsing $dSi(t_{n-1},t_n)$ and transactional $dFi(t_{n-1},t_n)$ components, with $t$ for events $e(t)$ varying in range from $t_{n-1}$ to $t_n$.

$$dS_i(t_{n-1},t_n) = \Sigma_k(H_k * \Sigma_{jct}(S_{ijkc}(t_n)*Q_{ijc}(t_n)*R_j(t_{n-1}))) / \Sigma_{jct}(Q_{ijc}(t_n)*R_j(t_{n-1})))/\Sigma_k(H_k)$$

$$dF_i(t_{n-1},t_n) = \Sigma_k(H_k * \Sigma_{jct}(F_{ijkce}(t)*G_{ijce}(t)*R_j(t_{n-1}))) / \Sigma_{jct}(G_{ijce}(t)*R_j(t_{n-1})))/\Sigma_k(H_k)$$

In simplified form, when no aspects or categories are considered, increases of endorsing and transactional reputations can be simplified as follows.

$$dS_i(t_{n-1},t_n) = \Sigma_{jt}(S_{ij}(t_n)*Q_{ij}(t_n)*R_j(t_{n-1})) / \Sigma_{jt}(Q_{ij}(t_n)*R_j(t_{n-1}))$$

$$dF_i(t_{n-1},t_n) = \Sigma_{jt}(F_{ije}(t)*G_{ije}(t)*R_j(t_{n-1})) / \Sigma_{jct}(G_{ije}(t)*R_j(t_{n-1}))$$

In practical implementation, either endorsing or transactional reputation can be used. In case of using both, a blended increase of reputation may be computed with blending factors $S$ and $F$ for each of the two reputations, respectively.

$$dP_i(t_{n-1},t_n) = (S * dS_i(t_{n-1},t_n) + F * dF_i(t_{n-1},t_n)) / (S + F)$$

Differential reputation can be further normalized by a maximum absolute reputation increase per time step:

$$P_i(t_{n-1},t_n) = dP_i(t_{n-1},t_n) / MAX_i(ABS(dP_i(t_{n-1},t_n)))$$

Based on reputation earned in the previous period from $t_o$ to $t_{n-1}$, the new reputation for latest time $t_n$ can be computed by blending the previous value with the differential one.

$$R_i(t_n) = ((t_{n-1} - t_o) * R_i(t_{n-1}) + (t_n - t_{n-1}) * P_i(t_{n-1},t_n)) / (t_n - t_o)$$

As it has been discovered in experiments discussed further on, a linear computation of reputation applied to experimental communities results in a quite nonlinear distribution of reputation values in the community, where very few members have very high values, but the rest of the community have reputations equal to zero. To improve the distribution for practical purposes, non-negative logarithmic differential reputation can be computed as follows, so the $lP_i(t_{n-1},t_n)$ can be used instead of $dP_i(t_{n-1},t_n)$ in the two formulae above.

$$lP_i(t_{n-1},t_n) = SIGN(dP_i(t_{n-1},t_n)) * log_{10}(1+ABS(dP_i(t_{n-1},t_n)))$$

Our reputation evaluation framework can be modified to allow earned reputation decay more quickly or slowly. We can apply extra blending factors to the most recent time interval importance and to earlier time intervals when computing $R_i(t_n)$, so that previously earned reputation values can decay faster or slower after being amended with the latest differential reputation.

It is also possible to compute more fine-grained reputations specific to different aspects or categories, as we will show for transactional differential reputation below. Based on these ideas, more precise reputations $R_{ic}(t_n)$, $R_{ik}(t_n)$ and $R_{ikc}(t_n)$ can be computed within the community.

$$dF_{ic}(t_{n-1},t_n)=\Sigma_k(H_k*\Sigma_{jt}(F_{ijkce}(t)*G_{ijce}(t)*R_j(t_{n-1}))) / \Sigma_{jt}(G_{ijce}(t)*R_j(t_{n-1})))/\Sigma_k(H_k)$$

$$dF_{ik}(t_{n-1},t_n)=\Sigma_{jct}(F_{ijkce}(t)*G_{ijce}(t)*R_j(t_{n-1})) / \Sigma_{jct}(G_{ijce}(t)*R_j(t_{n-1}))$$

$$dF_{ikc}(t_{n-1},t_n)=\Sigma_{jt}(F_{ijkce}(t)*G_{ijce}(t)*R_j(t_{n-1})) / \Sigma_{jt}(G_{ijce}(t)*R_j(t_{n-1}))$$

## 3   Design and implementation options

The computational framework suggested above can be designed and implemented in many possible ways, based on decisions made in respect to temporal scoping of the reputation calculation and its maintenance and storage options, as discussed further on. In the end, we introduce notions of "**Reputation consensus**", "**Proof-of-Reputation**" and "**Reputation mining**".

*Temporal scoping*
Performance of reputation system would depend on time scoping, based on interval spanning between cycles of reputation evaluations between times $t_{n-1}$ and $t_n$.

On one end, there is "**lifetime**" recalculation where all ratings between $t_0$ and $t_n$ are counted. In this case, it is possible to account for backdated changes in the ratings history to be accounted for with later re-calculation. However, this is much more expensive and time consuming. Also, in this case reputation decay can not be achieved as designed above and complication of differential reputation functions are required, introducing an extra time-bound weighting function which would give higher weights for more recent ratings.

On the other end, there is "**incremental**" recalculation with time intervals between $t_0$ and $t_n$ corresponding to intervals between subsequent transactions, so every transaction effects in global change of reputation. No reputation change delay may be experienced in this case yet implementation of this in a distributed way may get to be not trivial. At the same time, it might be beneficial for distributed systems not based on the blockchain.

In between the two above, there is "**up-to-date**" recalculation with time intervals between $t_0$ and $t_n$ being such as years, quarters, months, weeks, days etc. This would be more efficient and fast however reputations change may be delayed, getting outdated closer to the end of recalculation interval.

Finally, there is a hybrid between the last two such as "**blocked incremental**" recalculation where blocks of latest subsequent transactions are used to identify the time interval. It might be beneficial to have this implemented in distributed blockchain systems.

*Maintenance options*

From the maintenance perspective there are centralized, decentralized, and distributed options.

In the "**centralized**" implementation, all reputations are computed by a single party or system nominated and trusted to carry out this responsibility by community, called the Centralized Reputation Agency (CRA).

In the "**decentralized**" implementation, all reputations are computed by multiple parties or systems all trusted to carry out this responsibility by the community, called Decentralized Reputation Agencies (DRA). In such a case, there are two more options. One is that all agencies have to reach consensus on the current reputation state, so they may be called Coordinated Reputation Agencies (CDRA). The other option is that they all maintain reputations independently in which case they are Independent Reputation Agencies (IDRA).

In the "**distributed**" implementation, all members of the community are in charge of computing reputations, just as in blockchain, where all members in the community are in charge of logging and verifying transactions. In this case, it is also possible to have either a coordinated consensus of reputation states across all members, or for everyone to keep their own personal view of reputations of others.

*Storage options*

Within any options of the above, a few approaches of storing reputation data are possible, namely having it transient, persistent locally, or persistent globally in a decentralized or distributed way.

In the "**transient**" case, all reputations are always computed "on-the-fly" with endorsing and transactional data available. In the "**locally persistent**" case, all computed reputations are stored in local store by dedicated Reputation Agency or any member of the system, with no need to synchronize reputation status data with the others. In the "**globally persistent**" case, reputation data is maintained in decentralized or distributed way across all Reputation Agencies or all members of the system.

*Decentralized Reputation Consensus*

In the case of using the design from our earlier work [6], it is interesting to consider how multiple reputation agencies (RA) in CDRA reach decentralized consensus in respect to shared reputation data. The following administrative algorithm is proposed.

1. On each reputation calculation cycle, each RA submits the state of reputation data, providing reputations for all known members.

2. Within the cycle, each subsequent submitted state after the first one should be identical to the previous one.

3. If there is a submitted state not equal to previous, then the set of subsequent states is marked as being disputed and a warning to administration/monitoring services is sent.

4. Once the required system-wide minimum of identical states is received, the state is marked as valid and no more states are accepted.

5. Once the specified system-wide maximum of non-identical states is received and there is at least the minimum number of identical states, the state which is supported by most of submissions is marked as valid, no more states are accepted, and warning to administration/monitoring services is sent blaming non-identical state submitters.

6. If neither #4 nor #5 happens within specified system-wide time period starting the first submission, the consensus is considered broken, no reputation is updated, and a warning to administration/monitoring services is sent so entire system of CDRA has to be inspected.

7. In case the process above is implemented in distributed system implementing Proof-of-Reputation consensus algorithm, as defined below, voting for the accepted reputation state can be performed with account for reputation of the Reputation Agencies themselves, with appropriate changes in dispute resolution at step 5 above.

*Proof-of-Reputation*
In the case when the community implementing reputation calculations is performing distributed computing with the need for distributed consensus on the blockchain, the computed reputation may be used for the purpose of consensus achievement. The amount of reputation earned by a blockchain node impacts its chance to impact block formation in the same way that the amount of computing power or value of stake impacts on it in case of Proof-of-Work or Proof-of-Stake, respectively. This kind of consensus can be called as Proof-of-Reputation.

*Reputation Mining*
If reputations are being maintained in a distributed implementation, being computed through the participation of every member of the community, then it can be considered as a "mining" process. For this to happen, reputation states submitted by members must comply with the Decentralized Reputation Consensus protocol described above. In such case, those members who submit consistent reputation states accepted by the rest of members, should earn compensation on behalf of the community, as it happens in the Proof-of-Work consensus algorithm for solving cryptographic puzzles.

## 4 Approbation in social networks and blockchains

In an attempt to study practical computations of reputation in existing social networks and blockchains, practical studies have been carried out within Aigents social comput-ing platform [7,9], using social networks such as Facebook and Google+, and social networks based on blockchain, such as Steemit and Ethereum blockchain. We summarize some of our results below.

*Facebook and Google+*

For these social networks, using Aigents platform, it is possible to assess reputations of close social environments and a system users, based on interactions between the user and the social peers on its home page. Results from studies of a limited audience of users who agreed to share their experience display substantial reliability when communications are dense enough. On the other hand, if most communications of the user take place in groups or on home pages of the other users, reliability of results become insufficient. None of these results can be given here due to privacy restrictions.

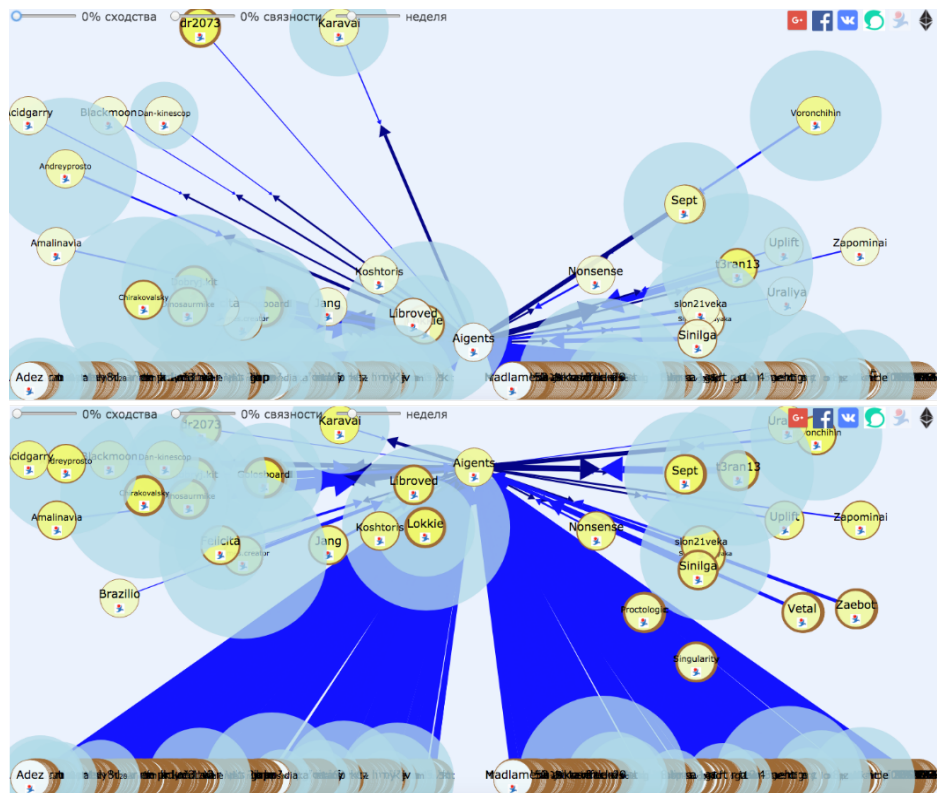

**Fig. 2.** Non-logarithmic (above) and Logarithmic (below) Reputation on transactional explicit and implicit ratings in the Steemit social network on the Steemit blockchain. The vertical position of a graph node corresponds to the reputation level of the member; the size of the "halo" circle around node represents similarity between respective member and member in the graph center; thickness of arrows between nodes corresponds to intensity of explicit and implicit ranking interactions between members.

*Steemit*

Since the Steemit social network is based on blockchain, all its results are publicly available for study. Computation of reputations on network-wide social graphs extracted from blockchain were studied for multiple public accounts. In this case, explicit transactional ratings such as votes on posts and comments, as well as implicit ratings such as indirect comments were used in simplified form. That is, no financial values in Steemit have been associated with ratings and implicit values of a comment are assumed positive regardless of comment sentiment. Studies have shown this simplified form to be in good correspondence with expected reality, according to an inspection of the history of interactions of real members. Along the way, it has been found that use of logarithmic reputations makes social structure more clearly identifiable, as shown on Fig.2.

*Ethereum*

In the Ethereum blockchain, only transactional financial ratings are available, so we used them to compute reputations of the members of the network. It has been found that distribution of values of financial transaction values is too non-linear, resulting in non-comprehensible distributions of reputation values. So we have had to involve logarithmic scaling of per-transaction values of financial transactions to get reasonable results for sensible reputation graphs.

## 5 Conclusion

We have come up with exhaustive model of computing reputations in multi-agent system based on different kinds of historical data representing various ways of interactions between agents. We have also suggested different ways to build reputation systems suitable for specific circumstances and requirements. We have also implemented part of the suggested design this with the Aigents social computing platform and made it available for public use at https://aigents.com web site. In the future, it is expected and recommended that more work be performed with respect to simulation modeling studying the resistance of the reputation system and reputation consensus against different sorts of reputation gaming and attack vectors attempting to takeover the system consensus. We also expect further development of distributed systems and reputation agency services based on the discussed principles and to improve them further.